\documentclass[review]{elsarticle}
\usepackage[breaklinks]{hyperref}

\makeatletter
\def\ps@pprintTitle{%
    \let\@oddhead\@empty
    \let\@evenhead\@empty
    \def\@oddfoot{\footnotesize\itshape
         {This is a post-print of an article published in Knowledge-Based Systems with doi:
         \href{https://doi.org/10.1016/j.knosys.2022.108453} {\color {blue} 10.1016/j.knosys.2022.108453}.}}%
    \let\@evenfoot\@oddfoot
    }
\makeatother

\usepackage{lineno}
\journal{Knowledge-Based Systems}
\usepackage{times,latexsym}
\usepackage{url}
\usepackage[T1]{fontenc}
\usepackage{amsmath}
\usepackage{caption}
\usepackage{graphicx}
\usepackage{stfloats}
\usepackage{subcaption}
\usepackage{xcolor}
\hypersetup{pdfauthor={Name}}

\DeclareSymbolFont{extraup}{U}{zavm}{m}{n}
\DeclareMathSymbol{\varheart}{\mathalpha}{extraup}{86}
\DeclareMathSymbol{\vardiamond}{\mathalpha}{extraup}{87}

\newcommand\upstrut{\rule{0pt}{12pt}}

\begin{document}
\begin{frontmatter}
\title{Local and Global Context-Based Pairwise Models for Sentence Ordering}
\author{Ruskin Raj Manku}
\author{Aditya Jyoti Paul\corref{mycorrespondingauthor}}
\cortext[mycorrespondingauthor]{Corresponding author\\
\indent\indent \textit{Email : \{ruskinraj\_manku, aditya\_jyoti\}@carlresearch.org}\\
\indent\indent Code available \href{https://github.com/RuskinManku/PairwiseModels4SO}{here}.}
\address{Cognitive Applications Research Lab, India}

\begin{abstract}
Sentence Ordering refers to the task of rearranging a set of sentences into the
appropriate coherent order. For this task, most previous approaches have
explored global context-based end-to-end methods using Sequence Generation
techniques. In this paper, we put forward a set of robust local and global
context-based pairwise ordering strategies, leveraging which our prediction
strategies outperform all previous works in this domain. Our proposed encoding
method utilizes the paragraph's rich global contextual information to predict
the pairwise order using novel transformer architectures. Analysis of the two
proposed decoding strategies helps better explain error propagation in pairwise
models. This approach is the most accurate pure pairwise model and our encoding
strategy also significantly improves the performance of other recent approaches
that use pairwise models, including the previous state-of-the-art, demonstrating
the research novelty and generalizability of this work.  Additionally, we show
how the pre-training task for ALBERT helps it to significantly outperform BERT,
despite having considerably lesser parameters. The extensive experimental
results, architectural analysis and ablation studies demonstrate the
effectiveness and superiority of the proposed models compared to the previous
state-of-the-art, besides providing a much better understanding of the
functioning of pairwise models.
\end{abstract}


\end{frontmatter}


\section{Introduction}
Understanding the underlying coherence structure in text is an important problem
in both Natural Language Processing and Natural Language Understanding, as
evidenced by several downstream tasks like multi-document summarization
\cite{barzilay2002inferring,nallapati2016summarunner}, retrieval-based question
answering \cite{SUN2007511,yu2018qanet,verberne2011retrieval}, concept-to-text
generation \cite{konstas2012concept}, events in storytelling
\cite{fan2019strategies,hu2020makes}, cooking steps in recipe generation
\cite{chandu2019storyboarding} and determining position of new sentences in
update summarization \cite{prabhumoye2019towards}.

The goal of Sentence Ordering \cite{barzilay2008modeling} is to arrange an
unordered set of sentences in the most logically coherent order. Coherent text
is expected to follow discourse properties like topical relevance, chronological
sequence and cause-effect \cite{hume1751philosophical,okazaki2004improving}.
Models trained for this task are especially helpful for automatic evaluation of
texts generated by language models and automatic scoring of essays
\cite{burstein2010using,miltsakaki2004evaluation,iter2020pretraining}.

In this paper, we dive into the strategies that can be used for getting robust
pairwise ordering models. These strategies can further be incorporated into
other methods that use pairwise models in any way, to improve the results. Many
previous research works have mentioned the possible disadvantages of pairwise
models, in terms of error propagation and their attempt to predict the relative
order of two sentences without any information of the global context of the
paragraph. We analyse this error propagation using two different decoding
strategies, and also incorporate global-context while predicting the pairwise
order, to improve accuracy of such models.

The main contributions of this paper can be summarized as follows:
\begin{itemize}
\item We propose novel local-context based pairwise ordering models
based on ensembles of BERT and ALBERT.
\item We present methods to incorporate the global context of a paragraph while
making pairwise predictions and demonstrate significant improvements in pairwise
ordering.
\item We put forward two different beam search based decoding mechanisms to
analyze the error propagation in pairwise models and show how one mechanism
supersedes another based on the target evaluation metric.
\item We empirically demonstrate that the proposed models outperform all
previous pure pairwise methods for three common benchmark datasets for all the
evaluation metrics.
\item We also showcase the generalizability of the proposed pairwise ordering
strategies; by incorporating the same in the previous state-of-the-art model,
and achieving the new state-of-the-art for the task.
\end{itemize}

\section{Related Works}

Early approaches proposed for this task relied on manually crafted linguistic
features by domain experts. \citet{lapata2003probabilistic} developed linguistic
vectors for sentences to model transition probabilities of adjacent sentences.
\citet{barzilay2004catching} captured domain specific content structures of text
in terms of topics the text addresses which are further represented as the
states of Hidden Markov Model \cite{1165342}. \citet{barzilay2008modeling}
developed entity-grid models to capture entity transitions between adjacent
sentences.  However, as these methods rely on hand-crafted features, they suffer
greatly when transferred to another domain or language.

Recently, various neural network based approaches have been proposed for this
task. Window networks \cite{li2014model} take as input a fixed window of
sentences and outputs the coherence probability of the window. The embeddings of
sentences are learned using Recurrent Neural Networks. \citet{chen2016neural}
proposed this task as a pairwise ranking task, an LSTM \cite{hochreiter1997long}
based model is used to determine the order of a pair of sentences and then
further beam search is used to predict the sub-optimal order. Pairwise
approaches were further explored in \citet{agrawal2016sort} and
\citet{li2016neural}. Almost all the methods proposed further frame this task as
a \textsl{seq2seq} task and apply end-to-end global context-based approaches.

\newpage
Language Models play an enormous role in various NLP tasks and are of utmost
importance in text coherence \cite{liu2020evaluating} and sentence ordering as
well. Most of the sentence ordering models till now have made use of BERT
\cite{devlin2018bert}, an excellent review of which was provided by
\cite{rogers2020primer}. Various extensions of BERT for different tasks have
also been proposed like by \citet{ZHAO2021107220} and \citet{SONG2021107408}.
ALBERT \cite{lan2020albert}, which was built with sentence ordering as a goal,
has been proposed to perform better when the problem is framed as a constraint
ordering problem in  \citet{prabhumoye2020topological} as well. A new
hierarchical transformer based approach for contextual sentence level
representation was proposed in \cite{lee2020slm}, which would also be useful for
sentence ordering tasks.

\citet{gong2016end} and \citet{logeswaran2018sentence} capture sentence
embeddings using LSTMs, they further capture the paragraph embedding using LSTMs
built on top of sentence embeddings and feed this as the initial state to a
pointer-network \cite{vinyals2015pointer} based decoder.  Self-attention
mechanism \cite{vaswani2017attention} based paragraph encoding is used in
\citet{cui-etal-2018-deep} to obtain a richer paragraph embedding compared to
LSTMs. This framework was further enhanced in \cite{wang2019hierarchical}, where
a self-attention mechanism based encoder is used for sentence-level embedding,
paragraph-level embeddings and the pointer network is replaced by self-attention
mechanism based decoder. Graph Recurrent Networks are built on entities in
\citet{yin2019graph} to capture the paragraph embedding. The problem is framed
as a ranking problem in \citet{kumar2020deep} and ranking losses like listMLE
\cite{xia2008listwise} are used to train the network. In \citet{zhu2021bert4so},
all the sentences are concatenated and fed to BERT along with special tokens,
these token representations are used as embeddings for sentence and listMLE is
used for scoring. \citet{9364604} enhance pointer network with conditional
sentence representation.

Recently there has been a surge in the use of pairwise models.
\citet{yin2020enhancing} enhances the performance of the pointer network decoder
significantly using pairwise ordering scores. In  \citet{cui-etal-2020-bert}, a
similar strategy is proposed, where they leverage BERT for pairwise level
predictions and improve the state of the art for the task. The same authors have
now come up with another approach in \citet{9444816}, where they diffuse BERT
for richer sentence representations. Some other methods do not use pointer
networks, but instead explore this problem with other solving strategies.
\citet{prabhumoye2020topological} framed the task of Sentence Ordering as a
constraint learning problem, where the constraint to be learned is the pairwise
ordering score of two sentences. To predict the optimal order, they represent
each sentence as a node in a graph, and a directed edge exists between two nodes
represented by $s_1$ and $s_2$ (from $s_1$ to $s_2$ ) if the pairwise score
indicates that $s_2$ comes after $s_1$ . The final order is then the topological
sort of this graph. \citet{Zhu_Zhou_Nie_Liu_Dou_2021} also created a similar
constraint graph and used it to get individual sentence embeddings using graph
neural networks \cite{zhou2020graph, zonghan2021gnn}. '

Considering this recent interest and advances in Sentence Ordering using pairwise models, this paper serves as a more fundamental exposition of the same; towards better understanding and improvement of key algorithmic and implementation aspects like incorporation of global context from other sentences and error propagation analysis for different decoding strategies.

\section{Methodology}
The goal of this paper is sentence ordering, that is to predict the optimal order $s_1^*,s_2^*,..,s_N^*$ given a set of $N$ sentences. In this work, we propose various models that aim to predict the pairwise ordering between sentences, and then leverage beam search on top of those predictions to search for the sub-optimal order.
\subsection{Pairwise Models}
\label{pairwise-models}
Given an out-of-order paragraph of $N$ sentences $P=[s_1,s_2,..s_N]$, the task
is to predict the pairwise order constraint between all the $N*(N-1)$ ordered
pairs. To this end, we propose the following models for this task.

\subsubsection{BERTPair}
BERT is a transformer mechanism based Language Model that attempts to learn
meaningful representations for the input tokens. Pre-training of BERT is
achieved using 2 tasks: Masked Language Modelling(MLM) and Next Sentence
Prediction(NSP). The first task helps it to learn meaningful intra-sentence
representations and the second task helps it to learn meaningful inter-sentence
representations. In order to find the pairwise constraint between $\mathtt{s_i}$
and $\mathtt{s_j}$ $\mathtt{(1<=i,j<=N, i \neq j)}$, the pair is fed in the form
the $\mathtt{[CLS]s_i[SEP]s_j[SEP]}$, which is similar to how the Next Sentence
Prediction task was trained, with this we are able to learn more meaningful
representations as compared to feeding each individual sentences as
$\mathtt{[CLS]s_i [SEP]}$ and $\mathtt{[CLS]s_j[SEP]}$. The classification
process can then be described with the following equations:

\vspace{-1.5em}
\begin{align}
\small
  \begin{split}
      & \mathtt{h_{BERT}=BERT([CLS]s_i[SEP]s_j[SEP])} \\
      & \mathtt{output=softmax(W_1(h_{BERT}))} \\
  \end{split}
\end{align}

Here, $\mathtt{W_1(d_{BERT}*2)}$ is a linear layer (weight matrix + bias)
and $\mathtt{d_{BERT}}$ represents the output embedding size of BERT. We can further define $\mathtt{pair\_score}$ as follows:
\begin{gather}
\label{eqpairscore}
\small
  \begin{split}
       & \mathtt{P(s_i<s_j|h_{BERT})=output[0], P(s_i>s_j|h_{BERT})=output[1]} \\
       & \mathtt{pair\_score=P(s_i<s_j|h_{BERT})} 
  \end{split}
\end{gather}
Here, $\mathtt{P(s_i<s_j)}$ represents the probability of $i^{th}$ sentence coming before the $j^{th}$ one, and $\mathtt{P(s_i>s_j)}$ represents vice-versa. This
model is same as the one proposed in \citet{prabhumoye2020topological}, where topological sort has been used for order prediction, however
we re-implemented it as we wanted to explore two different beam search based decoding
mechanisms, this gives us the flexibility to exploit the pairwise predictions in two different ways as described in Section \ref{decoding-mechanisms}. Hence, in our implementation, we can analyze error propagation and explore whether the same pairwise model can give different results for the Sentence Ordering metrics.

\subsubsection{ALBERTPair} ALBERT was proposed as an improvement to BERT. It
uses transformer mechanism like BERT, but with two architectural improvements,
$1)$ Factorized Embedding Parameterization and $2)$ Cross Layer Parameter
Sharing, these improvements reduce the number of parameters in ALBERT by a
significant number.($120M$ parameters in BERT base compared to $11M$ in ALBERT base
). However, the main motivation behind proposing this model for the pair order
prediction task is because of a different pre-training task than BERT. The
pre-training of ALBERT is achieved using 2 tasks, MLM(same as BERT) and Sentence
Order Prediction(SOP). SOP attempts to predict the correct order between two
consecutive segments. The pair order prediction task at hand is not exactly
similar to SOP, as we also attempt to predict the correct order between
sentences(segments) that are not necessarily consecutive, however we still
expect ALBERT to outperform BERT as pre-training significantly affects the
representations learned. The equations for classification can be described as
follows:

\vspace{-1em}
\begin{align}
\small
    \begin{split}
        & \mathtt{h_{ALBERT}=ALBERT([CLS]s_i[SEP]s_j[SEP])} \\
        & \mathtt{output=softmax(W_1(h_{ALBERT}))} \\
    \end{split}
\end{align}

Here,  $\mathtt{W_1(d_{ALBERT}*2)}$ is a linear layer and
$\mathtt{d_{ALBERT}}$ represents the output embedding size of ALBERT.

\subsubsection{EnsemblePair}
Ensembling of multiple models with techniques like bagging, stacking and
boosting have shown great promise in producing better results as compared to the
individual models in the ensemble. To this end, we propose an ensemble of BERT
and ALBERT to predict the pairwise ordering of two sentences. The variation in
representations learned by both models can be exploited to produce a robust local
context-based pairwise ordering model. The equations for classification can be
described as follows:

\vspace{-1em}
\begin{align}
\small
    \begin{split}
        & \mathtt{h_{ALBERT}=ALBERT([CLS]s_i[SEP]s_j[SEP])} \\
        & \mathtt{h_{BERT}=BERT([CLS]s_i[SEP]s_j[SEP])} \\
        & \mathtt{out\_logits=W_1(h_{ALBERT})+W_2(h_{BERT})} \\
        & \mathtt{output=softmax(out\_logits)}
    \end{split}
\end{align}

Here,$\mathtt{W_1(d_{ALBERT}*2)}$ and $\mathtt{W_2(d_{BERT}*2)}$ are 
linear layers; $\mathtt{d_{ALBERT}}$ and $\mathtt{d_{BERT}}$ is the output
embedding size of ALBERT and BERT, respectively.

\subsubsection{GlobalPair} With this model, we attempt to incorporate the Global
context of a paragraph while predicting the pairwise ordering of two sentences.
To this end, we use the encoder of the transformer model as described by 
\citet{vaswani2017attention}, henceforth referred to as \textit{transformer\_encoder}. Each transformer block has a multi-head self
attention sub-layer and position wise feed-forward neural network sub-layer. Layer normalization
\cite{lei2016layer} is also used after each sub-layer, further details about this encoder architecture can be found in \cite{vaswani2017attention}.

\begin{figure}
\centering
\includegraphics[width=\textwidth]{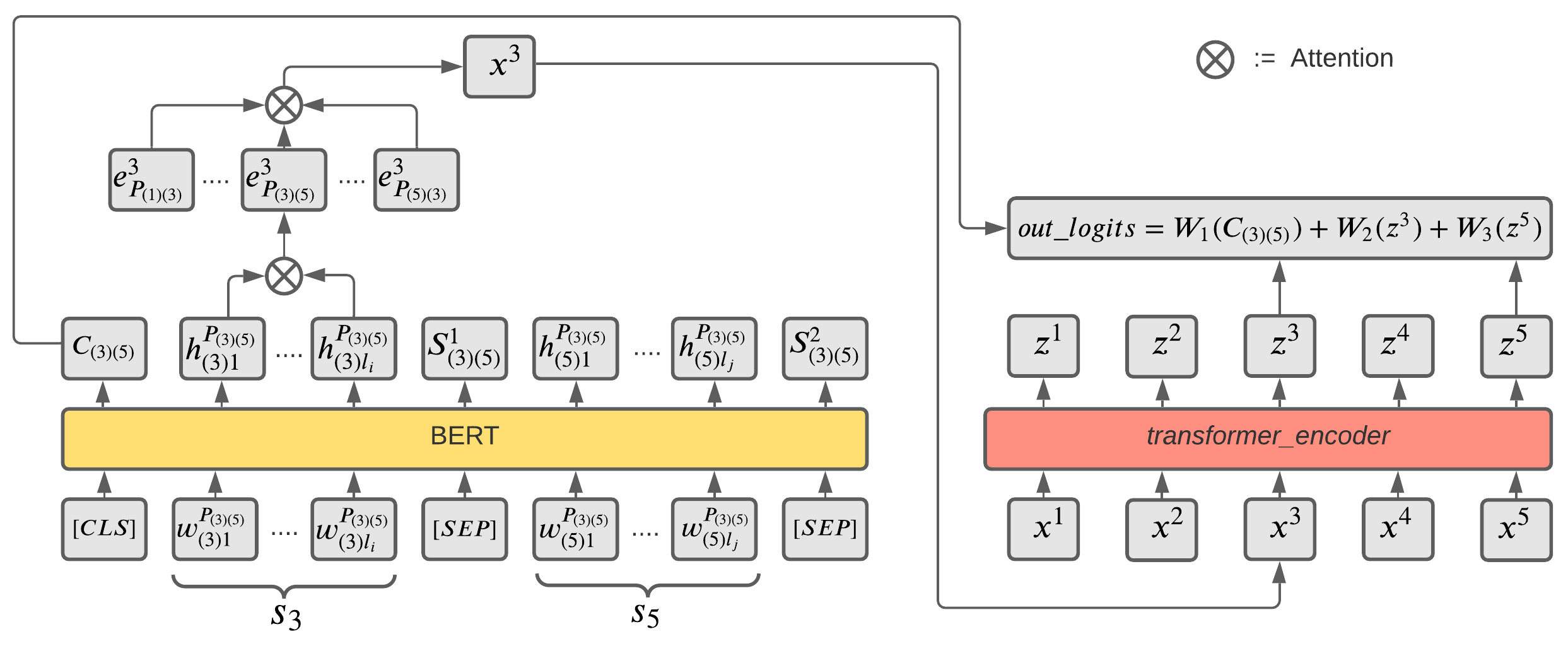}
\caption{Schematic of the \textit{GlobalPair} model. Here, we show the pairwise order prediction process for $s_3$ and $s_5$ in a paragraph with $N=5$ sentences. We get $e_{P_{(3)(5)}}^3$ by passing the pair $P_{(3)(5)}$ through BERT and then using attention on the hidden states of the output. Similarly through attention on all the $2(N-1)=8$ values $[e_{P_{(1)(3)}}^3$, $e_{P_{(2)(3)}}^3$,..., $e_{P_{(3)(4)}}^3$, $e_{P_{(3)(5)}}^3]$, we get the sentence representation $x^3$. Repeating this process to get sentence representations $x^i$ for all sentences and then inputting them to the \textit{transformer\_encoder} gives $z^i$ values. Finally, for predicting the order between $s_3$ and $s_5$, we use the hidden output of $[CLS]$ token i.e. $C_{(3)(5)}$, along with global-context rich vectors $z^3$, $z^5$ to obtain the final order.} 
\vspace{1.5em}
\label{fig:global}
\end{figure}

\vspace{-2em}
\sloppy Given an unordered paragraph $P = [s_1 , s_2 ....s_N ]$, we extract the
embedding of each sentence by building upon the Hierarchical Relational Sentence
Encoder proposed in \citet{cui-etal-2020-bert}. The schematic of this \textit{GlobalPair} is illustrated in Figure \ref{fig:global}. Formally, if we feed
sentence pair $s_i$ and $s_j$ to BERT, each containing $l_i$ and $l_j$ tokens
respectively, BERT transforms the sequence to $(C_{ij},h_{i1}^{P_{ij}}, ... ,
h_{il_i}^{P_{ij}},S_{ij}^1,h_{j1}^{P_{ij}}.....h_{jl_j}^{P_{ij}},S_{ij}^2)$.
Here, $C_{ij}$ is the final $[CLS]$ token embedding, $S_{ij}^1$ is the embedding
of $[SEP]$ token following $s_i$, $S_{ij}^2$ is the embedding of $[SEP]$ token
following $s_j$, $h_{i1}^{P_{ij}}...h_{il_i}^{P_{ij}}$ and
$h_{j1}^{P_{ij}}...h_{jl_j}^{P_{ij}}$ are the output token embeddings for $s_i$
and $s_j$ in pair $P_{ij}$, respectively. We can then obtain embedding
$e_{P_{ij}}^i$, the representation for $s_i$ in the $P_{ij}^{th}$ pair by using
attention mechanism across its hidden states
$h_{i1}^{P_{ij}}...h_{il_i}^{P_{ij}}$. The following equations describes this process,

\vspace{-1.5em}
\begin{align}
\label{attention_mechanism}
    \centering
    \begin{split}
        & \mathtt{u_{ik}=tanh(W_{w}(h_{ik}^{P_{ij}}))}
        \\
        & \mathtt{\alpha_{ik}=\frac{exp(v_w(u_{ik}))}{\sum_{k=1}^{l_i}exp(v_w(u_{ik}))}} \\
        &  \mathtt{e_{P_{ij}}^i=\sum_{k=1}^{l_i}\alpha_{ik}h_{ik}^{P_{ij}}} 
    \end{split}
\end{align}

Here $W_w$ and $v_w$ are learnable parameters. Each sentence is present in
$2(N-1)$ pairs, thus we will obtain $e_1^i,e_2^i,e_3^i....e_{2(N-1)}^i$
embedding vectors for $i^{th}$ sentence. We can obtain the embedding $x_i$ for
$i^{th}$ sentence by applying the same attention mechanism described in Equation
\eqref{attention_mechanism} to these $2(N-1)$ embeddings. We pass these
embeddings through a transformer to obtain the final global context rich
embedding $z_i$ for $i^{th}$ sentence.

\vspace{-2.5em}
\begin{align}
\small
    \centering
    \begin{split}
        & \mathtt{z_1,z_2,..,z_N=transformer\_encoder(x_1,x_2,..,x_N)} 
    \end{split}
\end{align}

The embeddings $z_i$ for $i$ in $(1,N)$ thus obtained, contain information about
the sentence along with information from other sentences(global information)
captured from the pairwise interaction and self-attention interaction. We
incorporate this information with the [CLS] token information. To predict the
pairwise ordering of two sentences $s_i$ and $s_j$ , we use the following
equations:

\vspace{-1.5em}
\begin{align}
\label{equation6}
    \small
    \begin{split}
        & \mathtt{h_{BERT}=BERT([CLS]s_i[SEP]s_j[SEP])} \\
        & \mathtt{out\_logits\!=\! W_1(h_{BERT}) \!+\! W_2(z_i) \!+\! W_3(z_j)} \\
        & \mathtt{output=softmax(out\_logits)} \\
    \end{split}
\end{align}

Here, $\mathtt{W_1(d_{BERT}*2)}$, $\mathtt{W_2(d_{BERT}*2)}$ and
$\mathtt{W_3(d_{BERT}*2)}$ are linear layers; $\mathtt{d_{BERT}}$ is the
output embedding size of BERT. We call this model BERT-GlobalPair subsequently,
as we can further replace BERT with ALBERT and their Ensemble as the pair
encoding model, we call these two models ALBERT-GlobalPair and
Ensemble-GlobalPair respectively.

\newpage
\subsection{Order prediction algorithms}
\label{decoding-mechanisms}
We propose two different beam-search based decoding algorithms to predict the
sub-optimal coherent order of a shuffled paragraph by leveraging the pairwise
predictions. As described before, we can get $\mathtt{pair\_score}$ from the $\mathtt{output}$ of above models as shown in Equation \eqref{eqpairscore}. 

Initially at $t=1$, each member of the beam consists of the individual sentences $s_1$ to $s_N$ with each having a score of $0$. At any step $t$ of the decoding phase, we create a new beam from the $(t-1)^{th}$ beam by adding sentences at the $t^{th}$ position of each candidate solution, finally we sort the new beam in descending order of the beam score, and consider the top $B$(beam size) members for the next step.. The schematic of the two decoding algorithms is shown in Figure \ref{fig:decoding}. Formally, at $t\neq1$, any member in the beam can be represented as $s^*_1 , s^*_2 , ...s^*_{t-1}$ , we add the new
sentence $s^*_t$ (not already present in the set $s^*_1 , .., s^*_{t-1}$ ) and calculate the new beam score as follows:
\begin{figure}
\centering
\includegraphics[width=\textwidth]{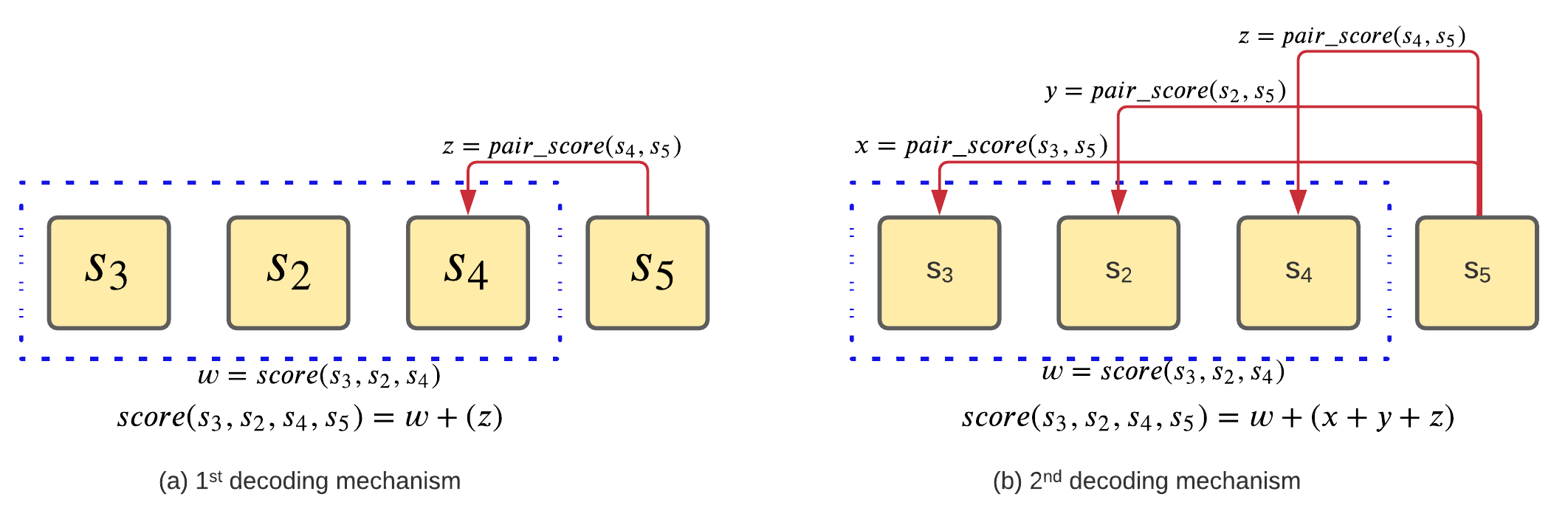}
\caption{Schematics of the two decoding mechanisms. Here we have a beam of size 3 containing sentences $s_3(s_1^*)$, $s_2(s_2^*)$ and $s_4(s_3^*)$ with a beam score of $w$. To calculate the new score in a scenario where $s_5$ is added to the beam, in (a) we only add the pair\_score of $s_5$ with the last sentence, i.e, $s_4$. In (b), we consider the pair\_score with all sentences in the beam.}
\label{fig:decoding}
\end{figure}
\begin{enumerate}
\item We only consider the pairwise score of $s^*_{t-1}$ and $s^*_t$ in
addition to the previous score of the beam. That is,
\begin{align}
            \small
              score(s^*_1,..,s^*_t)=score(s^*_1,..,s^*_{t-1})
              + pair\_score(s^*_{t-1},s^*_t)
\end{align}
\item We consider the total score of all the sentences in the candidate solution
with $s^*_t$ in addition to the previous score of the beam. That is,
\begin{align}
              score(s^*_1,..,s^*_t)=score(s^*_1,..,s^*_{t-1})
              + \sum_{i=1}^{t-1}pair\_score(s^*_i,s^*_t)
\end{align}
\end{enumerate}

In this paper, we have presented 6 strategies as discussed in Section
\ref{pairwise-models}, and the decoding strategy for each model is indicated by
its suffix, \textbf{-1} and \textbf{-2}. 

\section{Experiments}
This section describes the datasets used for evaluation, the evaluation metrics,
the baselines we compare our models to, training setup, the main results of our
experiments and the ablation studies.

\subsection{Datasets}
\begin{itemize}
\item \textbf{SIND captions} \cite{huang2016visual}: A visual story telling
dataset. It is one of the challenging benchmark datasets for this problem
because the sentences describe pictures in sequence and hence information is
missing without the pictures. There are 40155 training stories, 4990 validation
stories and 5055 testing stories. The dataset is highly balanced in the sense
that each story is composed of 5 sentences. This dataset is openly available.
\item \textbf{ROCStories} \cite{mostafazadeh-etal-2016-corpus}: A commonsense
story dataset and hence the paragraphs are very coherent. There are a total of
98162 stories, with each story containing exactly 5 sentences. Following the
previous works, we split this dataset using a 8:1:1 random split to get 78529
stories in the train set, 9816 stories in the validation set and 9817 stories in
the test set. This dataset is openly available.
\item \textbf{ACL abstracts} \cite{logeswaran2018sentence}: This dataset is
obtained by parsing abstracts from ACL Anthology Network(AAN) corpus. All the
extracts of papers published upto 2010 are used for training, year 2011 for
validation and years 2012-2013 for testing. There are 8569 train abstracts, 962
validation abstracts and 2626 test abstracts.
\end{itemize}

\subsection{Evaluation metrics}
\begin{itemize}

\item \textbf{Accuracy(Acc)} :This measures the percentage of sentences in a
paragraph for which the absolute position was correctly predicted. For a corpus
$D$ with $N$ paragraphs, this can be formulated as
$Acc=\frac{1}{N}\sum_{i=1}^{i=N}\frac{1}{n_i}\sum_{j=1}^{j=n_i}I\{e_j^i=o_j^i\}$,
where $I$ is the indicator function, $n_i$ is the number of sentences in
$i^{th}$ paragraph, $e_j^i$ is the $j^{th}$ sentence in the predicted order and
$o_j^i$ is the $j^{th}$ sentence in correct order for that paragraph.

\item \textbf{Kendall's tau($\tau$)} : This measures the relative ordering
distance between the predicted order and the correct order \cite{lapata2006automatic} and for one paragraph
can be formulated as $\tau=1-2*\frac{\#inversions}{\binom{n}{2}}$, where $n$ is
the number of sentences in paragraph and $\#inversions$ is the number of pairs
in the predicted sequence with incorrect relative order. The value for this
metric ranges from $-1$ to $+1$.

\item \textbf{Perfect Match Ratio(PMR)} : This measures the ratio of paragraphs
in the corpus for which the predicted order exactly matches the correct order.
$PMR=\frac{1}{N}\sum_{i=1}^{i=N}I\{e^i=o^i\}$, where I denotes the indicator
function, $e^i$ is the predicted order and $o^i$ is the correct order.
\end{itemize}

\vspace{1em}
\subsection{Baselines}
We can divide all the previous proposed methods for this task into two separate
categories. The first category includes pure pairwise models, i.e. models that
use pairwise ordering as a primary encoding mechanism. It consists of the
following:
\begin{itemize}
\itemsep-0.3em 
\item \textbf{Seq2Seq}      \cite{li2016neural}, 
\item \textbf{B-TSort}     \cite{prabhumoye2020topological}, and
\item \textbf{Pairwise+GNN} \cite{Zhu_Zhou_Nie_Liu_Dou_2021}.
\end{itemize}

In the second category, are methods that do not leverage pairwise orderings as a
primary encoding or decoding mechanism. This contains methods that use ListMLE
for decoding, or Pointer Networks. This category consists of the following
models, that we use as baselines:
\begin{itemize}
\itemsep-0.3em 
\item \textbf{LSTM+PtrNet}      \cite{gong2016end}, 
\item \textbf{LSTM+Seq2Seq}     \cite{logeswaran2018sentence},
\item \textbf{ATTOrderNet}      \cite{cui-etal-2018-deep}, 
\item \textbf{HAN}              \cite{wang2019hierarchical}, 
\item \textbf{SE-Graph}         \cite{yin2019graph},
\item \textbf{RankTxNet}        \cite{kumar2020deep}, 
\item \textbf{PtrNet+pairwise}  \cite{yin2020enhancing}, 
\item \textbf{BERSON}          \cite{cui-etal-2020-bert},
\item \textbf{PtrNet+Conditional}   \cite{9364604},
\item \textbf{ERSON}            \cite{9444816} and
\item \textbf{BERT4SO}          \cite{zhu2021bert4so}.
\end{itemize}

Note that \textbf{PtrNet+pairwise} and \textbf{BERSON} use pairwise ordering as
a way to enhance the performance of the pointer network, and not as a primary
encoding mechanism. For \textbf{RankTxNet} multiple models have been proposed,
of which we have reported the best score achieved by any model. For
\textbf{B-TSort}, we have utilized the openly available code to get the results
for ROCStories dataset.

\subsection{Training Setup}
All the models are implemented in Pytorch \cite{NEURIPS2019_9015}. We use the
HuggingFace library \cite{Wolf2019HuggingFacesTS} to train BERT and ALBERT
models. Both models are initialized with the weights of their "base" versions
for fair comparison, and ensemble compatibility. The transformer network in GlobalPair model is made up of 2
transformer blocks, with hidden size of 768, feed-forward intermediate layer
size 3072 and 12 attention heads. We use Adam optimizer and a batch size of $2$
for all three datasets. For SIND and ROCStories, learning rate used is
$5*10^{-6}$ and decay factor is $0.2$ per epoch. For the ACL dataset, we found the training
to be quite stochastic, the variance was found to be high for global context models and thus initial weights can have a pronounced effect on final results. Hence, we encourage the readers to train the models multiple times to achieve optimal results using the learning rate of $1*10^{-5}$, decayed to $1*10^{-6}$ after $1^{st}$ epoch and
constant afterwards. A beam size of $64$ is used for decoding.

\tabcolsep=0.15cm
\begin{table*}[t]
    \resizebox{\textwidth}{!}{
    \begin{tabular}{c|ccc|ccc|ccc}
        \hline\upstrut
        Models                     & \multicolumn{3}{c}{SIND} & \multicolumn{3}{c}{ROCStories} & \multicolumn{3}{c}{ACL}                                                                                                             \\
                                   & Acc                      & $\tau$                         & PMR                               & Acc             & $\tau$          & PMR             & Acc             & $\tau$          & PMR             \\
        \hline\upstrut
        Seq2Seq $^\diamondsuit$ & - & 0.19 & 12.50 & - & 0.34 & 17.93 & - & - & - \\
        B-TSort(ACL 2020) $^\diamondsuit$         & 52.23                    & 0.60                           & 20.32                             & 72.85           & 0.80            & 50.02           & 69.22           & 0.83            & 50.76           \\
        Pairwise+GNN (AAAI 2021)$^\diamondsuit$ \hspace{-0.6em} & - & 0.58 & 19.07 & - & 0.81 & 49.52 & - & 0.82 & 49.81 \\
        \hline\upstrut
        BERTPair-1 $^\varheart$                & 52.41*                   & 0.590                          & 21.99*                            & 73.03*          & 0.808           & 51.13*          & 71.31           & 0.813           & 49.13*          \\
        BERTPair-2 $^\varheart$                & 52.34                    & 0.602*                         & 20.25                             & 72.82           & 0.812*          & 49.89           & 71.57*          & 0.821*          & 48.25           \\
        \hline\upstrut
        ALBERTPair-1 $^\varheart$              & 53.80                    & 0.611                          & 24.17*                            & 74.53*          & 0.821           & 53.80           & 74.33*          & 0.833           & 53.54*          \\
        ALBERTPair-2 $^\varheart$              & 54.16*                   & 0.623*                         & 23.18                             & 74.42           & 0.823*          & 53.10           & 74.31           & 0.835*          & 53.19           \\
        \hline\upstrut
        EnsemblePair-1 $^\varheart$            & 54.38                    & 0.616                          & 25.22*                  & 75.57* & 0.829           & 55.60* & 75.64* & 0.840           & 54.79* \\
        EnsemblePair-2 $^\varheart$            & 54.65*        & 0.629*                & 24.12                             & 75.37           & 0.830* & 54.59           & 75.58           & 0.844* & 54.43           \\
        \hline\upstrut
        BERT-GlobalPair-1 $^\varheart$              & 56.14                    & 0.631                          & 26.15*                            & 79.44*          & 0.859           & 60.96*          & 75.71*          & 0.841           & 54.90*          \\
        BERT-GlobalPair-2 $^\varheart$              & 55.24*                   & 0.639*                         & 23.12                             & 79.19           & 0.865*          & 59.69          & 75.48           & 0.842*          & 54.66           \\
        \hline\upstrut
        ALBERT-GlobalPair-1 $^\varheart$              & 56.04                    & 0.632                          & 27.03*                            & 79.50*          & 0.860           & 61.11*          & 75.74*          & 0.841           & \textbf{55.01}*          \\
        ALBERT-GlobalPair-2 $^\varheart$              & 56.36*                   & 0.650*                         & 25.63                             & 79.25           & 0.868*          & 59.72         & 75.48           & 0.850*          & 54.69           \\
        \hline\upstrut

        Ensemble-GlobalPair-1 $^\varheart$          & \textbf{56.36}*                   & 0.632                         & \textbf{27.20}*                           & \textbf{79.53}*           & 0.860          & \textbf{61.25}*          & \textbf{75.76}*           & 0.842          & 55.00           \\

        Ensemble-GlobalPair-2 $^\varheart$              & 56.11                   & \textbf{0.653}*                         & 25.71                             & 79.26           & \textbf{0.871}*          & 59.88          & 75.50           & \textbf{0.851}*          & 54.71           \\
        \hline
    \end{tabular}} 
    \caption{Results on the sentence ordering task for pure
    pairwise models. $\diamondsuit$ indicates previously reported scores.
    $\varheart$ indicates models implemented in this paper. * represents the
    best result out of the two decoding mechanisms. Numbers in \textbf{bold}
    represent the best result obtained across all the models}
    \vspace{-1.5em}
\label{big-table}
\end{table*}
\subsection{Main results}
\vspace{-0.5em}

\tabcolsep=0.23cm
\begin{table*} [!b]
\centering
    \resizebox{0.85\textwidth}{!}{%
    \begin{tabular}{c|cc|cc|cc}
        \hline
        Model        & \multicolumn{2}{c}{SIND} & \multicolumn{2}{c}{ROCStories} & \multicolumn{2}{c}{ACL abstracts}                                           \\
                            & P-Acc                    & std                            & P-Acc                             & std          & P-Acc   & std          \\
        \hline
        BERTPair            & 79.79\%                  & $\pm$16.03\%                   & 90.13\%                           & $\pm$11.56\% & 90.43\% & $\pm$11.76\% \\
        ALBERTPair          & 80.78\%                  & $\pm$16.05\%                   & 90.81\%                           & $\pm$11.48\% & 91.45\% & $\pm$11.52\% \\
        EnsemblePair        & 81.12\%                  & $\pm$15.90\%                   & 91.22\%                           & $\pm$11.10\% & 91.65\% & $\pm$11.46\% \\
        BERT-GlobalPair     & 81.51\%                  & $\pm$15.96\%                   & 92.84\%                           & $\pm$10.26\% & 91.73\% & $\pm$11.44\% \\
        ALBERT-GlobalPair   & 82.16\%                  & $\pm$16.10\%                   & 92.90\%                           & $\pm$10.30\% & 91.79\% & $\pm$11.40\% \\
        Ensemble-GlobalPair & 82.31\%                  & $\pm$16.12\%                   & 92.95\%                           & $\pm$10.24\% & 91.81\% & $\pm$11.42\% \\
    \end{tabular}}
    \caption{Pairwise Classification results for our 6 proposed models. P-Acc is
    the mean accuracy of all the pairs for each paragraph in the dataset. std is the
    standard deviation of the pairwise accuracy}
\vspace{-1em}
\label{pairwise-results}
\end{table*}

\subsubsection{Pure Pairwise Strategies}
Table \ref{big-table} shows the performance of our models compared against other
pairwise approaches i.e. our baselines of the first category. Table
\ref{pairwise-results} shows the pairwise classification results for the models.
It shows that \textbf{the pre-training task of ALBERT helps it outperform BERT
on all the datasets by a considerable margin}.

Another important observation is that \textbf{a small increase in the
classification accuracy and a small decrease in the standard deviation can
translate to considerable gains in the evaluation metrics, especially for PMR}.
Many end-to-end methods have mentioned the issue of error propagation in pair-
wise models,we proposed two different decoding mechanisms to observe that. For
the $1^{st}$ decoding mechanism, we are relying lesser on the output of pairwise
ordering model as compared to the $2^{nd}$ model: For finding the score of a
beam of size $n$ , we are referring the pairwise score $n-1$ times in the former
as compared to $\binom{n}{2}$ times in the latter. Since PMR is a very strict
measure, i.e, a paragraph is not a perfect match with even just one sentence
misplaced, the more times we refer to the pairwise score, the more are the
chances of referring an incorrect pairwise prediction and the lesser the PMR
should be. This can be observed for all the models across all the datasets. The
difference in PMR is high for SIND dataset between the two decoding mechanisms,
but lesser for other two datasets. At the same time, $2^{nd}$ decoding mechanism
always outperforms $1^{st}$ decoding mechanism for $\tau$, with a considerable
margin, due to the $2^{nd}$ strategy and scoring metric lining up well with each other.

From Table \ref{big-table}, it can be observed that a local-context based ensemble of ALBERT and BERT is
able to take advantage of two very different sentence representations learned by
different pre-training tasks and is able to outperform the individual models in
the Ensemble, i.e BERT and ALBERT, across both decoding strategies.
\vspace{-6pt}

\subsubsection{Improving any Network that uses Pairwise Methods (e.g. BERSON)}

We implemented BERSON so as to experiment with the pairwise strategies presented
in this paper. (AL)BERSON is a variant of BERSON, where we replace BERT
with ALBERT as pair encoding mechanism and in (Ensemble)BERSON we use
their Ensemble for pair encoding. We do not change anything else in the
architecture of BERSON. Note that we do not report scores for a
(Global)BERSON model, because in BERSON, the pairwise representation is
concatenated with a global-information rich sentence embedding in the pointer
network decoding stage, and hence no improvements are observed with that model.
\\
\vspace{-6pt}

\tabcolsep=0.15cm
\begin{table*}[tp]
    \resizebox{\textwidth}{!}{%
    \begin{tabular}{c|ccc|ccc|ccc}
        \hline
        Models                     & \multicolumn{3}{c}{SIND} & \multicolumn{3}{c}{ROCStories} & \multicolumn{3}{c}{ACL abstracts}                                                                                                             \\
                                   & Acc                      & $\tau$                         & PMR                               & Acc             & $\tau$          & PMR             & Acc             & $\tau$          & PMR             \\
        \hline\upstrut
        LSTM+PtrNet $^\diamondsuit$              & -                        & 0.48                           & 12.34                             & -               & 0.71            & 40.44           & 58.20           & 0.69            & -               \\
        LSTM+Seq2Seq(AAAI 2018) $^\diamondsuit$     & -                        & 0.49                           & 13.80                             & -               & 0.71            & 35.81           & 58.06           & 0.73            & -               \\
        ATTOrderNet(EMNLP 2018) $^\diamondsuit$   & -                        & 0.49                           & 14.01                             & -               & -               & -               & 63.24           & 0.73            & -               \\
        HAN(AAAI 2019) $^\diamondsuit$          & -                        & 0.50                           & 15.01                             & -               & 0.73            & 39.62           & -               & 0.69            & 31.29           \\
        SE-Graph(IJCAI-19) $^\diamondsuit$        & -                        & 0.52                           & 16.22                             & -               & -               & -               & 64.64           & 0.78            & -               \\
        RankTxNet(AAAI 2020) $^\diamondsuit$     & -                        & 0.56                           & 15.59                             & -               & 0.76            & 38.02           & -               & 0.77            & 39.18           \\
        PtrNet+pairwise(AAAI 2020) $^\diamondsuit$   & -                        & 0.53                           & 17.37                             & -               & 0.76            & 46.00           & -               & -               & -               \\
        BERSON(EMNLP 2020) $^\heartsuit$ & 57.88 & 0.632 & 29.91 & 81.53 & 0.864 & 65.99 & 73.49 & 0.812 & 52.20 \\ 
        PtrNet+Conditional(ICSC 2021) $^\diamondsuit$ & - & 0.567 & 17.46 & - & 0.772 & 42.01 & - & - & - \\
        ERSON(TPAMI 2021) $^\heartsuit$ & 56.81 & 0.639 & 28.82 & 82.76 & 0.875 & 67.12 & 73.98 & 0.823 & 54.50 \\
        BERT4SO $^\diamondsuit$ & - & 0.591 & 19.07 & - & 0.848 & 55.65 & - & 0.807 & 45.41 \\
        \hline\upstrut

        (AL)BERSON $^\varheart$ & 59.65 & 0.650 & 32.55 & 82.61 & 0.881 & 68.48 & 75.15 & 0.831 & 55.44  \\
        (Ensemble)BERSON $^\varheart$ & \textbf{59.67} & \textbf{0.661} & \textbf{33.02} & \textbf{82.91} & \textbf{0.887} & \textbf{69.02} & \textbf{75.41} & \textbf{0.849} & \textbf{56.03} \\
    \end{tabular}}
    
    \caption{Results on the sentence ordering task for methods that do not use
    pairwise ordering as primary encoding strategy. $\diamondsuit$ indicates
    previously reported scores. $\heartsuit$ indicates models implemented by
    ourselves, and hence result can vary from numbers reported in original
    paper. $\varheart$ indicates models implemented in this paper. Numbers in
    \textbf{bold} represent the best result obtained across all the models}
    \label{ptrnet-results}
    \vspace{-1.3em}

\end{table*}

\newpage Table \ref{ptrnet-results} shows the results thus obtained as compared to
previous methods that use Pointer Network or an end-to-end strategy, i.e. the
baselines of the second category. As evident, there is a strong increase in
performance of BERSON. This is a result of the fact that, we are now getting
much better pairwise ordering representations, which indeed translate into a
better performance for the pointer network as well. In Table \ref{p-acc-berson},
we note the pairwise accuracy of the classifier in BERSON and its proposed
improvements to show that an \textbf{increase in accuracy with our pairwise
strategies translates to a more robust pointer network and improved
predictions}.

\tabcolsep=0.11cm
\begin{table}[bp]
\begin{small}
    \begin{center}
        \begin{tabular}{c|c|c|c}
            Model/Dataset    & SIND & ROCStories & ACL                           \\
            \hline
            BERSON       & 78.92\% & 88.28\% & 88.63\% \\
            (AL)BERSON & 79.82\%  & 89.20\%  & 89.15\% \\
            (Ensemble)BERSON & 79.96\% & 89.31\% & 89.29\% \\
        \end{tabular}
    \end{center}
\end{small}
\vspace{-1.5em}
\caption{Pairwise Accuracy of BERSON and its variants}
\label{p-acc-berson}
\end{table}

Based on our observations, these improvements are generalizable and can be
extended to other models that use pairwise strategies too, like those proposed
by \citet{prabhumoye2020topological}, \citet{Zhu_Zhou_Nie_Liu_Dou_2021} and
\citet{9444816}. These models leverage BERT to get a pairwise prediction; simply
replacing BERT with the other pairwise strategies proposed in this paper will
directly improve results. These gains are especially significant for
B-TSort and Pairwise+GNN, as they leverage pairwise
predictions directly to form constraint graphs. 

\subsection{Discussion}
In this section, we discuss certain interesting and insightful observations from
this paper and also cover some further analyses and ablation studies.

\subsubsection{Rate of Improvement in Pure Pairwise Models}
Firstly, we showcased how our vanilla BERTPair approach beats the previous
constraint graph based approach despite them using pairwise models to  enhance
their performance. Although it was already expected that ALBERT would give
performance improvements over BERT, to the best of our knowledge, we're the
first to systematically analyze the performance and improvements of the same.
Then we present a novel EnsemblePair strategy that improves upon both BERTPair
and ALBERTPair.

\begin{figure}[b!]
\centering
\includegraphics[scale = 0.47]{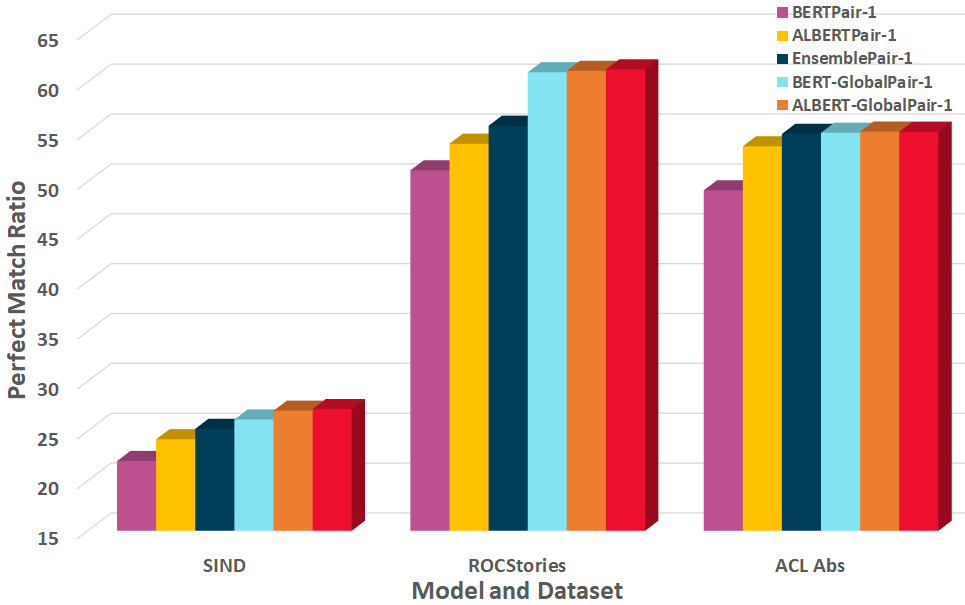}
\caption{Rate of PMR evolution across the 6 models for each dataset using the
$1^{st}$ decoding strategy, y-axis values start from 15.}
\label{pmr-evolution}
\end{figure}

All our proposed strategies we discussed above, however, did not use any form of
global paragraph context. To remedy this shortcoming and to make our models more
robust in the light of global context, we imbibe global information into each of
these three models and introduce BERT-GlobalPair, ALBERT-GlobalPair and
Ensemble-GlobalPair. As observed in the local models, performance improves as we
move from BERT-GlobalPair to ALBERT-GlobalPair to ultimately
Ensemble-GlobalPair.

\begin{figure}[b!]
\centering
\includegraphics[scale = 0.47]{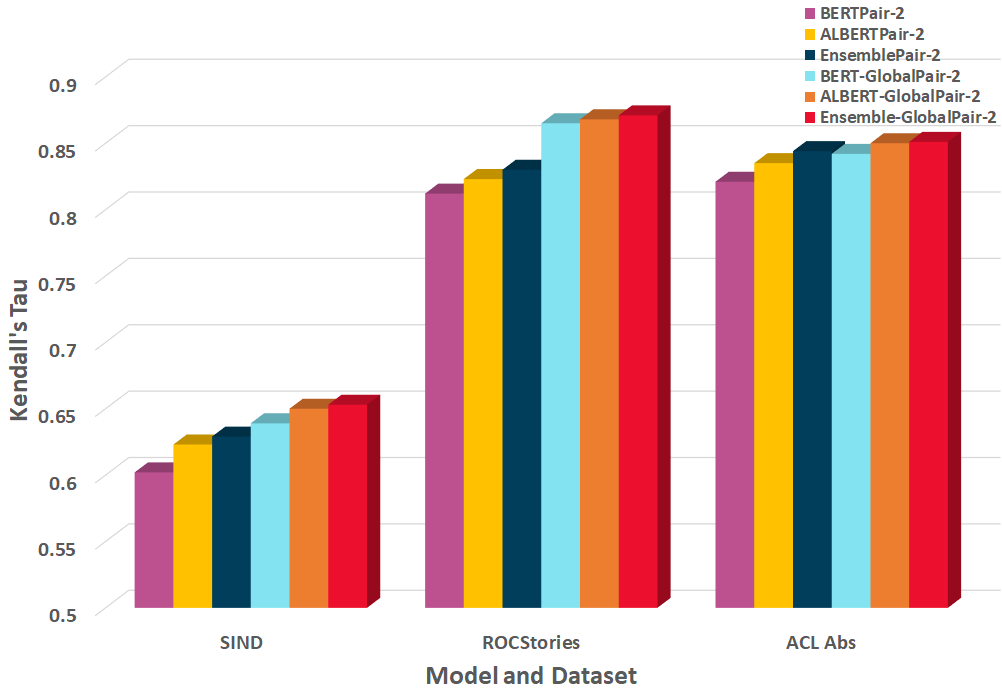}
\caption{Rate of Kendall's Tau evolution for the 6 models for each dataset using
the $2^{nd}$ decoding strategy, y-axis values start from 0.5.}
\label{kt-evolution}
\end{figure}

\textbf{Adding the global-context of a paragraph while making pairwise
predictions boosts the accuracy of the classifier, which in turn improves the
results for the evaluation metrics of the sentence ordering task}. While making
pairwise prediction, we are able to add the global information relevant to that
sentence with the help of self-attention transformer mechanism. This gain is
significant across all datasets, especially ROCStories. We note however, that
changing  BERT to ALBERT or their Ensemble in GlobalPair does not improve the
results as significantly as the gain is when switching from local-context based
BERTPair to ALBERTPair or EnsemblePair. This can be observed from the PMR and
Kendall's Tau evolution in Figures \ref{pmr-evolution} and \ref{kt-evolution}
respectively.

\subsubsection{Ablation: GlobalPair with single sentence representations}
In our proposed GlobalPair method, we extract sentence level embeddings by
adding a transformer layer on top of embeddings obtained from individual pairs.
However, it's possible to extract sentence embeddings directly by passing the
sentence through BERT, in a $[CLS]s_i[SEP]$ token form, using the $[CLS]$ token
embedding as the sentence embedding, and then passing the embeddings thus
obtained through a transformer to obtain global rich embedding for each
sentence. More formally, the classification process in this case can be
described as follows.
\begin{align}
\small
    \centering
    \begin{split}
        & \mathtt{x_i=BERT([CLS]s_i[SEP]) \forall i\in(1,N)} \\
        &  \mathtt{z_1,z_2,..,z_N=transformer(x_1,x_2,..,x_N)} 
    \end{split}
\end{align}

\begin{table}[bp]
\tabcolsep=0.11cm
\begin{small}
    \begin{center}
        \begin{tabular}{c|cc|cc|cc}
            Model/Dataset    & \multicolumn{2}{c}{SIND} & \multicolumn{2}{c}{ROCStories} &  \multicolumn{2}{c}{ACL} \\
                       & P-Acc                           & std                              & P-Acc   & std     & P-Acc & std     \\
            \hline
            Single-GlobalPair       & 80.85\% & $\pm$16.14\% & 91.16\% & $\pm$11.26\% & 91.52\% & $\pm$11.47\% \\
            BERT-GlobalPair & 81.51\% & $\pm$15.96\% & 92.84\% & $\pm$10.26\% & 91.73\% & $\pm$11.44\% \\
        \end{tabular}
    \end{center}
\end{small}
\caption{Comparing pairwise accuracy of Single-GlobalPair, where we feed single sentences to Ensemble Model, with BERT-GlobalPair.}
\label{single-table}
\end{table}

The embeddings $z_1,z_2...z_N$ thus obtained, contain global rich information
and can be used as in Equation \eqref{equation6}. In Table \ref{single-table},
we compare BERT-GlobalPair with the above model, but instead of using BERT for
getting representation from $[CLS]$ token, we use Ensemble of BERT and ALBERT,
to get the best possible representation.

BERT-GlobalPair outperforms this model for all datasets. The performance of this
model is very close to that of EnsemblePair, which suggests that global
information exchange using single sentence representations is not beneficial
from single sentence representations. This suggests that language models like
BERT give quite better representations when the sentences are fed as a pair.

\section{Conclusion}

Pairwise models have been shown to be useful in various NLP and NLU tasks in the
past, including sentence ordering. Due to multiple reported shortcomings of
these pairwise models in terms of error propagation and their inability to
account for the paragraph context, research focus shifted towards end-to-end
models like pointer networks. 

The primary novelty of our work is to demonstrate that it was not the inherent
shortcoming of the pairwise model class of approaches but the way they had been
implemented in the past which resulted in their sub-optimal performance. We have
presented 6 pure pairwise models, each with 2 decoding strategies and
demonstrate their performance to be far superior to all previous pairwise models
and almost at par with the previous state-of-the-art pointer network based
approaches. 

We emphasize that it is our state-of-the-art strategies which make pairwise
methodologies so robust and efficient and that this has been the missing puzzle
piece for this class of problems all these years, because none of the prior art,
to the best of our knowledge, used global context information at the time of
pairwise classification. This can be clearly observed by the enormous
performance jump from EnsemblePair without global information to even the simple
BERT-GlobalPair infused with global information.

We present some deeper insights into how pairwise models work; most notably, our
first decoding strategy which involves much lesser error propagation due to
fewer sentence pair comparisons, performs better for PMR which is a strict
evaluation metric, on the other hand, for Kendall Tau, which is inversely
proportional to the number of pairwise incorrect flips, the second decoding
strategy considering all the sentence pairs gives a higher score.

We also implement (AL)BERSON and (Ensemble)BERSON, to showcase how our pairwise
strategies are so novel and generalizable that it can be used to improve the
previous SOTA approach BERSON, also beating the latest model ERSON.

Thus, in this paper, we present a set of novel strategies that make pairwise
models extremely powerful in various text coherence tasks like sentence
ordering, empirically prove its superiority to all previous pairwise approaches
and demonstrate its generalizability by using it to improve the previous
state-of-the-art, surpassing the performance of all prior works in this domain. 

\section{Future Work}
We show that GlobalPair improves the pairwise accuracy of ordering. In future,
it would be fruitful to explore other strategies to incorporate Global context
in the paragraph. Currently, global information fed is independent of the
sentence ordering within the paragraph. However, information regarding the
position of other sentences could be useful for making pairwise predictions,
this can be achieved by adding positional embeddings in transformers and
training for multiple permutations. 

Another possible area of exploration is to have specific pre-training strategies
for Sentence Ordering. As we show that pre-training of ALBERT helps it perform
better, even more relevant pre-training strategies might allow the models to
achieve significant improvements for this task. Exploration of different
Ensembling strategies for BERT and ALBERT like assigning novel parameter based
weights can also be delved into, which would further improve the performance of
pairwise ordering and in turn the larger set of text coherence problems.

\bibliography{references}
\end{document}